\documentclass[a4paper]{article}

\usepackage[english]{babel}
\usepackage[utf8x]{inputenc}
\usepackage[T1]{fontenc}

\usepackage[a4paper,top=3cm,bottom=2cm,left=3cm,right=3cm,marginparwidth=1.75cm]{geometry}
\usepackage{setspace}
\usepackage[table]{xcolor}
\usepackage{amsmath}
\usepackage{graphicx}
\usepackage[colorinlistoftodos]{todonotes}
\usepackage[colorlinks=true, allcolors=blue]{hyperref}
\usepackage{wrapfig}
\usepackage{subcaption}
\usepackage{natbib}
\usepackage{xcolor}
\usepackage{amsfonts}

\usepackage[linesnumbered,boxruled,lined]{algorithm2e}
\usepackage{algpseudocode}
\usepackage{booktabs}

\newcommand{\X}{\mathbf{X}}
\newcommand{\z}{\mathbf{z}}

\newcommand{\Y}{\mathbf{Y}}
\newcommand{\R}{\mathbb{R}}

\newcommand*\samethanks[1][\value{footnote}]{\footnotemark[#1]}

\author{Liam Welsh\thanks{Corresponding author:~\href{mailto:liam.welsh@mail.utoronto.ca}{liam.welsh@mail.utoronto.ca}} $^{,}$\thanks{Authors contributed equally}
\\ Department of Statistical Sciences,\\ University of Toronto
\\
\and
Phillip Shreeves\samethanks[2]
\\ Department of Statistics and Actuarial Science,\\ Simon Fraser University
\\
}

\title{A Non-Parametric Bootstrap for Spectral Clustering}

\begin{document}
\maketitle

\begin{abstract}

Finite mixture modelling is a popular method in the field of clustering and is beneficial largely due to its soft cluster membership probabilities. A common method for fitting finite mixture models is to employ spectral clustering, which can utilize the expectation-maximization (EM) algorithm. However, the EM algorithm falls victim to a number of issues, including convergence to sub-optimal solutions. We address this issue by developing two novel algorithms that incorporate the spectral decomposition of the data matrix and a non-parametric bootstrap sampling scheme. Simulations display the validity of our algorithms and demonstrate not only their flexibility, but also their computational efficiency and ability to avoid poor solutions when compared to other clustering algorithms for estimating finite mixture models. Our techniques are more consistent in their convergence when compared to other bootstrapped algorithms that fit finite mixture models.\end{abstract} 

\vspace{.5cm}

\newpage
\section{Introduction} \label{intro}

As the world becomes more saturated with big data, the ability to produce interpretable statistical models from high dimensional data is a task of great significance. In the field of clustering, unsupervised learning algorithms can be used to find patterns or groups within data and construct models in these high dimensional settings  (see~\cite{bouveyron2014model,celebi2016unsupervised}). These models can take the form of a finite mixture of distributions, a probabilistic model that can be used to represent sub-populations within data~\cite{mclachlan2019finite}. A main issue with constructing models, such as finite mixture models (FMMs) for high dimensional data is the fitting of these models and parameter estimations can be incredibly slow, converge to local solutions, or lack interpretation. Even in low dimensional settings, clustering algorithms may be fraught with issues, including converging to poor local maxima solutions or overfitting to the data available which results in the optimal log-likelihood but having overconfident group membership estimates. We demonstrate this in Section~\ref{sec:motex}. \cite{andrews2018addressing} has demonstrated and addressed the issue of overfitting and sub-optimal convergence in FMMs through the use of a non-parametric bootstrap. FMMs are of particular interest as they provide a probabilistic result for group memberships, allowing for valuable insight into analysis and model interpretation; see~\cite{mclachlan2019finite}. However, fitting a FMM in a high dimensional setting is computationally expensive, as it requires repeated inversions of $p \times p$ covariance matrices, where $p$ is equal to the number of variables. Further, the bootstrapping procedure~\cite{andrews2018addressing} employs is itself computationally non-trivial. To address the issues of overfitting and sub-optimal convergence in high dimensional FMMs, we propose the combination of a common dimensionality reduction clustering utilized in spectral clustering and the non-parametric bootstrap. Spectral clustering is a versatile clustering technique that is useful for high dimensional settings; see~\cite{vempala2004spectral} and~\cite{von2007tutorial}. Despite its versatility, it is still prone to the aforementioned issues.


We develop two novel algorithms and show that our proposed algorithms not only address the issue of overfitting and convergence to sub-optimal solutions, but are also significantly faster than the current bootstrapped approaches. The remainder of this paper is organized as follows. A motivating example is introduced in Section~\ref{sec:motex}. Section~\ref{background} follows with the relevant background information required to understand the methodological contribution of this research, which is presented in Section~\ref{method}. Our work includes two bootstrap-augmented algorithms for spectral clustering and a convergence criterion. Algorithm performances are then analyzed using simulated and real-world data sets in Section~\ref{sec:results} and compared against prior algorithms. Concluding remarks are provided in Section~\ref{conc}.

\section{Motivating Example}\label{sec:motex}

\begin{figure}
     \centering
     \begin{subfigure}[t]{0.49\textwidth}
         \centering
         \includegraphics[width=\textwidth]{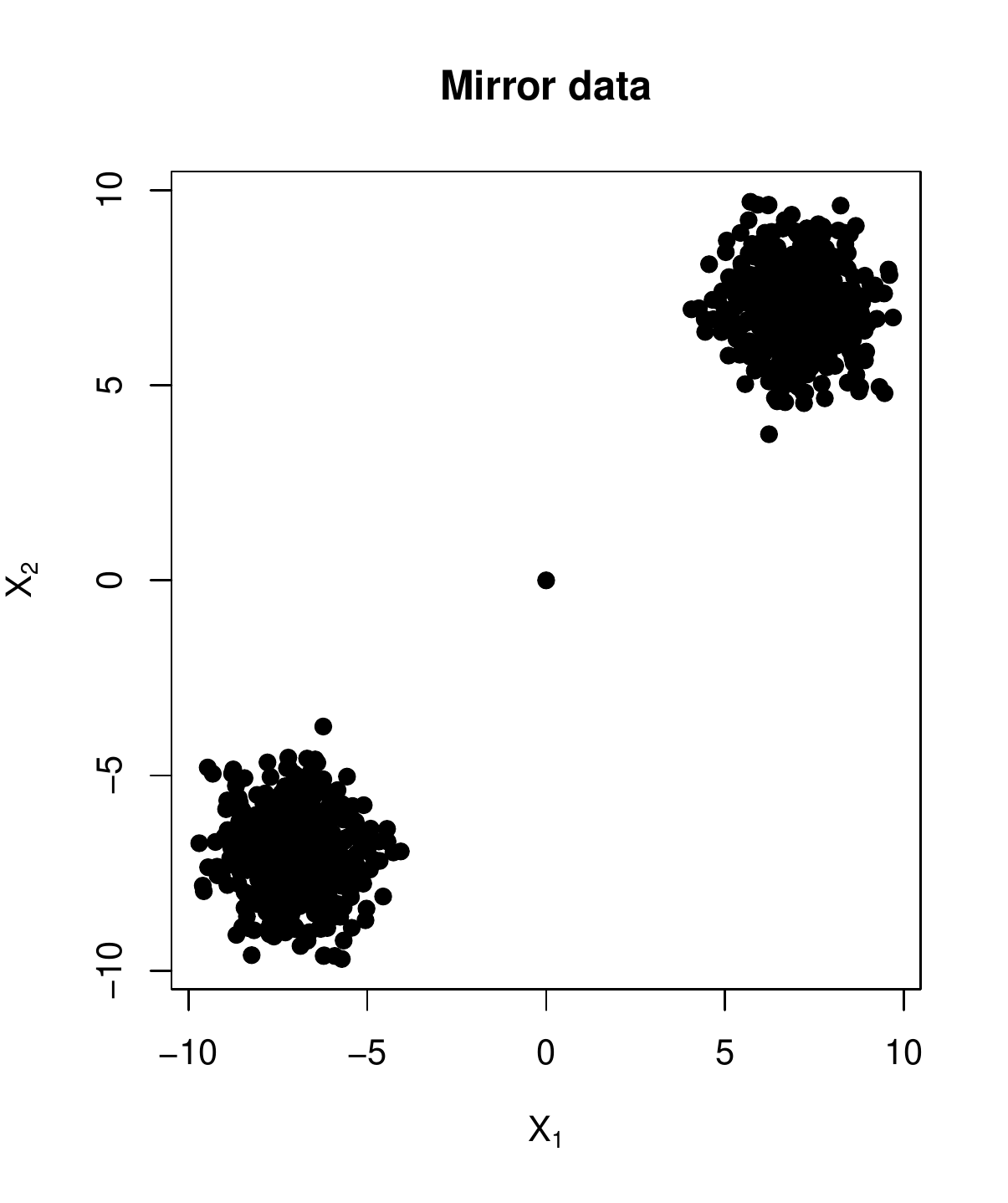}
         \caption{First two variables of mirror data. This is two of the 150 total variables available. However, pairwise plots of any two variables will look similar.}
         \label{fig:mirror}
     \end{subfigure}
     \hfill
     \begin{subfigure}[t]{0.49\textwidth}
         \centering
         \includegraphics[width=\textwidth]{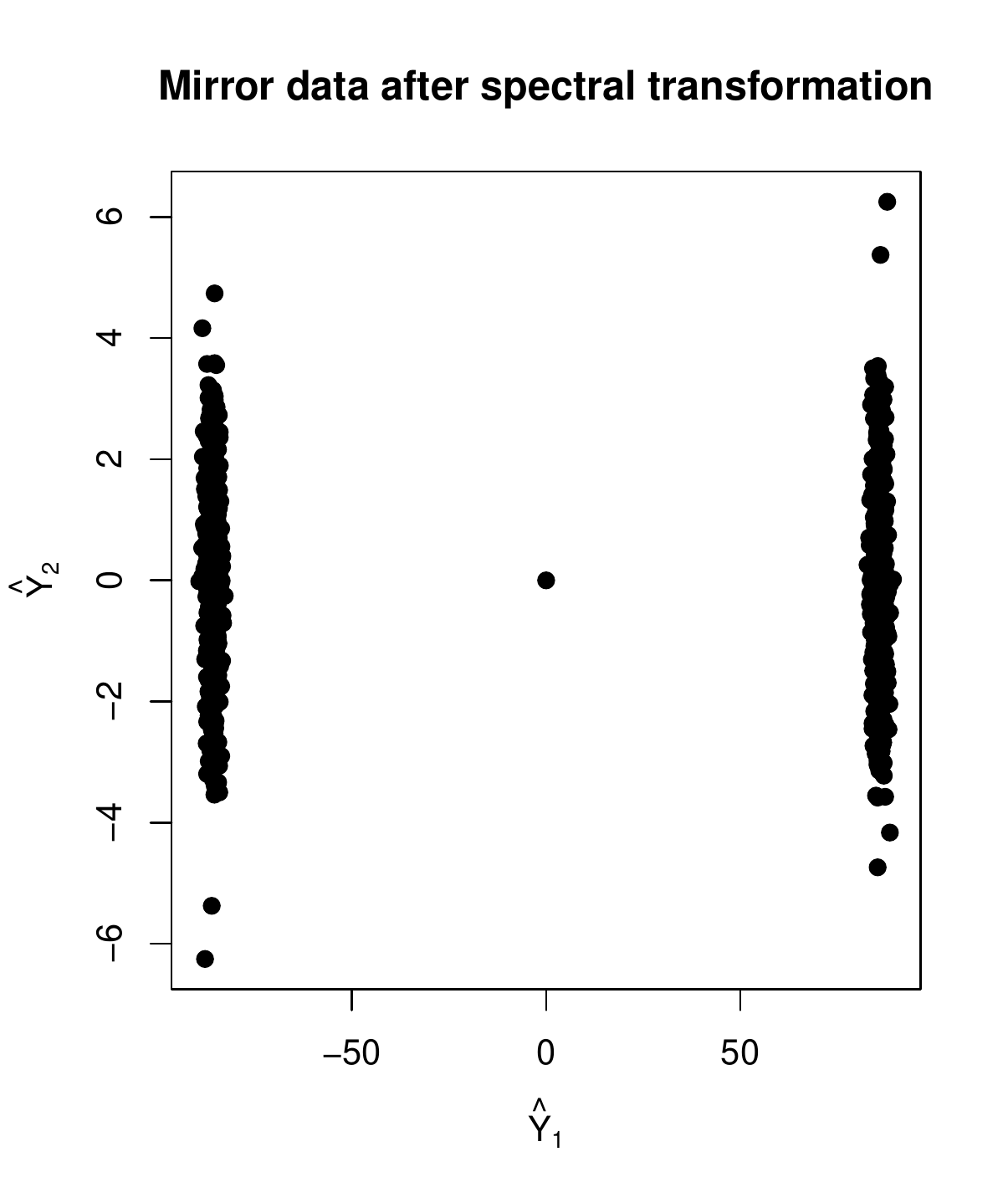}
         \caption{Remaining two components of mirror data after spectral transformation $\hat \Y = \X\,V^{'}_2$ where $V^{'}_2$ is the first two columns of the unitary matrix of $\X$ from the SVD $X = U\,\Sigma\, V^{'}$.}
         \label{fig:transformedmirror}
     \end{subfigure}
        \caption{Original (left) and transformed (right) mirror data.}
        \label{fig:mirror_both}
\end{figure}

We extend the motivating example of~\cite{andrews2018addressing} from a two-dimensional setting to a high dimensional setting. Specifically, we construct a sample of 500 data points with 150 variables centered at $(7,\dots,7)$ with standard normal noise. We then take a copy of this data reflecting it across the hyperplane $X = - Y,$ where $X,Y\in\R^{150}$, providing us with two distinct groups. Finally, one observation is created at $(0,\dots,0)$ so that it is equidistant from each group. The first two variables of this data set are plotted in Figure~\ref{fig:mirror}. While spectral clustering is well suited for noisy data, this extremely stylized example with well separated groups demonstrates the issue at hand.


Intuition leads one to believe that the probability of the centre point belonging in either cluster would be $50\%$ in a two group solution. However, this is not found to be the case with many clustering algorithms. The $k$-means algorithm simply labels the point as belonging to one group or another as it is a hard clustering algorithm that provides no statistical reference. Conversely, the benchmark expectation-maximization (EM) algorithm with unconstrained covariances heavily overfits and finds the probability of the centre point belonging to one group to be over $99.999\%$, and practically 0\% to belong the other. A constrained or equal covariance assumption in a high dimensional setting becomes increasingly strong, and difficult to verify given real data.

~\cite{andrews2018addressing} has demonstrated that finding a two group solution in the mirror data is the result of the EM algorithm converging to a global maxima. While this may be ideal with respect to the MLE, it is flawed from an inference perspective. This issue remains in the solution when a spectral transformation is used on the data, see Figure~\ref{fig:transformedmirror}. In particular, we employ the decomposition technique used in Algorithm 1 by~\cite{loffler2021optimality}. Using this methodology, we find that it estimates the centre point's group membership probabilities to be $(1,0)$, counter-intuitive to the point's relative location.

~\cite{andrews2018addressing} has shown that applying a non-parametric bootstrap can address this issue of overfitting in the EM algorithm. In high dimensions, the method of~\cite{andrews2018addressing} is extremely slow.~\cite{shreeves2019bootstrap} further expand upon this work by using the non-parametric bootstrap to augment the alternating expectation-conditional maximization (AECM) algorithm. This uses the assumption that the observations arise from a factor analysis model and was developed in a mixture model framework by \cite{ghahramani1996algorithm, mclachlan2000factor}. This work is further outlined in Section~\ref{BootAECM}. While this removes the need for the direct computation of multiple $p \times p$ inversions at each bootstrap iteration, it is still required once per iteration to find the likelihood of the averaged parameter space. Additionally a large number of matrix multiplications are needed to solve for the latent space, slowing the algorithm further. Our proposed methodology greatly reduces the space on which to perform matrix calculations, achieving significant gains in computational efficiency while converging to an appropriate solution.

\section{Background} \label{background}

\subsection{Finite Mixture Models} \label{finiteMMs}
FMMs~(\cite{mclachlan2019finite}) are commonly used for clustering to provide a statistical representation of an underlying group structure and have been used in a variety of applications; see \citep{mcnicholas2016mixture,torres2002language,yu2012online,georgiades2022trajectories}. A clear benefit of FMMs is that it is a true statistical model. Once fitted, an FMM can provide a probabilistic result when grouping new data; a feature not available with hard clustering techniques such as the $k$-means methods and hierarchical trees. As such, clustering using an FMM is known as a soft clustering technique. Mathematically, FMMs are a convex combination of $G$-many statistical densities, and can be expressed as
\begin{equation}\label{eq:FMM}
    f(\mathbf{x}_i) = \sum_{g=1}^G \pi_g\, f_g(\mathbf{x}_i \mid \Theta_g).
\end{equation}
Here, $\mathbf{x}_i$ is a $p$ dimensional observation from a data matrix $\X$ that is $n \times p$, $G$ is the number of underlying groups in the model, $\pi_g$ is the mixing proportion of group $g$ (with the constraint $\sum_{g=1}^G \pi_g = 1$), and $f_g$ is the underlying distribution of the $g^{th}$ group component with parameters $\Theta_g$. In this work, we direct our focus to mixtures of multivariate normal distributions, which are represented as
\begin{equation}\label{eq:GuassFMM}
    f(\mathbf{x}_i) = \sum_{g=1}^G \pi_g\, \phi(\mathbf{x}_i \mid \mu_g, \Sigma_g),
\end{equation}
where $\mu_g$ and $\Sigma_g$ are the mean vector and covariance matrix of the $g$-th group, respectively. FMMs are well studied for the multivariate normal distribution and other families of distributions~\citep{andrews2011model,franczak2013mixtures,punzo2016parsimonious}. 
Take $\mathbf{z}_i = (z_{i1}, \dots, z_{iG})$ to be the $G$-dimensional vector of group membership probabilities for $\mathbf{x}_i$, such that $\sum_{g=1}^G z_{ig} = 1$. Then $\pi_g$ can be viewed as the \textit{a priori} probability of observation $i$ belonging to group $g$, with an \textit{a posteriori} probability of
\begin{equation}
    \text{P}(z_{ig} = 1 \mid \mathbf{x}_i)= \frac{\pi_g\,\phi(\mathbf{x}_i \mid \mu_g, \Sigma_g)}{\sum_{j=1}^G \pi_j \,\phi(\mathbf{x}_i \mid \mu_j, \Sigma_j)}.
\end{equation}
This quantity can be estimated as
\begin{equation}\label{eq:ZEst}
    \hat{z}_{ig} = \frac{\hat{\pi}_g\, \phi(\mathbf{x}_i \mid \hat{\mu}_g, \hat{\Sigma}_g)}{\sum_{j=1}^G \hat{\pi}_j\, \phi(\mathbf{x}_i \mid \hat{\mu}_j, \hat{\Sigma}_j)},
\end{equation}
where $\hat{\mu}_g$, $\hat{\Sigma}_g$, and $\hat{\pi}_g$ for $g=1,...,G$ are the maximum likelihood estimates determined from the model likelihood
\begin{equation}
    \mathcal{L}(\Theta) = \prod_{i=1}^n\sum_{g=1}^G \pi_g \,\phi(\mathbf{x}_i \mid \mu_g, \Sigma_g).
\end{equation}
The conventional multivariate normal finite mixture model requires 
\begin{equation}\label{eq:freepar}
    \underbrace{G-1}_{\text{mixing proportions}} + \underbrace{G\,p}_{\text{group means}} + \underbrace{\frac{G\,p\,(p+1)}{2}}_{\text{group covariances}}
\end{equation}
free parameters, with the parameter contributions displayed below Equation~\eqref{eq:freepar}. The largest contributor arises from the estimation of the group covariance matrices, which is a natural target for parsimony~\cite{mcnicholas2016mixture}. A method to do so is discussed in Section~\ref{BootAECM}. However, multiple other techniques have been used to address this; see \cite{banfield1993model,celeux1995gaussian,tipping1999mixtures}. The most common way to estimate these parameters is in an iterative style, alternating between updating the maximum likelihood estimates and updating the posterior probabilities~\cite{dempster1977maximum}. This is known as the expectation-maximization (EM) algorithm and is further detailed in the following section.

\subsection{EM Algorithm} \label{EMAlg}
The EM algorithm~\citep{dempster1977maximum,mclachlan2007algorithm} is an extremely common method to estimate Gaussian FMMs. It iterates between two steps; the expectation (E) step and the maximization (M) step. The E-step calculates the estimated expected value of $Z_{ig}$ for all $i=1,...,n$ and $g=1,...,G$, found in~\eqref{eq:ZEst}, while the M-step calculates the maximum likelihood estimates of $\hat{\pi}_g$, $\hat{\mu}_g$, and $\hat{\Sigma}_g$. Updates for the maximum likelihood estimates are given as
\begin{equation}\label{eq:EMmle}
    \hat{\pi}_g = \frac{n_g}{n},\ \ \ \ \ \ \hat{\mu}_g = \frac{1}{n_g}\sum_{i=1}^n \hat{z}_{ig}\,\mathbf{x}_i,\ \ \text{\&} \ \ \ \hat{\Sigma}_g = \frac{1}{n_g}\sum_{i=1}^n \hat{z}_{ig}\,(\mathbf{x}_i - \hat\mu_g)(\mathbf{x}_i-\hat\mu_g)^\prime.
\end{equation}
An algorithm summary \citep{mcnicholas2016mixture} is provided in Algorithm~\ref{algo:EM}. A common convergence criterion used for the EM algorithm is 'lack of progress', which compares the updated version of the log-likelihood, denoted $\ell = \log\left(\mathcal{L}(\Theta)\right)$, against its previous iteration. The lack of progress criterion is given as
\begin{equation}
    \mid \ell^{(k)}-\ell^{(k-1)}\mid < \epsilon
\end{equation}
where $\epsilon > 0$. However, we note that there are other criterion available. Research regarding the `optimal convergence criterion' is outside the scope of this manuscript and we refer readers to~\cite{mcnicholas2016mixture} for further details. Further evaluation of model performance is often calculated using the Bayesian information criterion (BIC), such that
\begin{equation}
     \text{BIC} = 2\,\ell(\Theta) - \rho\cdot\text{log}(n),
\end{equation}
where $\rho$ is the number of free parameters in the model. BIC includes a penalization term for number of parameters relative to the number of observations.

\begin{algorithm}[htbp]
	\caption{EM algorithm for finite GMM estimation}\label{algo:EM}
    \footnotesize
	\KwIn{$\X$}
   Initialize $\hat{\z}$ using either k-means or discrete uniform randomization\;
   \While{not converged}{
    \For{$g = 1,\dots, G$}{
        Update the MLEs using Equations in~\eqref{eq:EMmle}\;
        Compute $\hat z_g$ using the new MLE estimates\;
        Check for convergence\;  
    }
    }
    \KwOut{$\hat {\pi}_g$, $\hat {\mu}_g$, and $\hat {\Sigma}_g$ for $g=1,\dots,G$}
\end{algorithm}

\subsection{Bootstrapping and the BootEM Algorithm} \label{BootEM}

The bootstrap procedure, introduced in~\cite{efron1981nonparametric} and~\cite{efron1982jackknife}, is a resampling technique that enables estimates of parameter standard errors that otherwise may not be possible to calculate. Specifically, for a parameter $\theta$ with estimate $\hat\theta$, we can calculate new estimates of $\hat\theta$ by sampling with replacement from the data $\X$, creating a (non-parametric) bootstrap sample. Repeating this process $M$ times, we can compute the standard deviation of the parameter estimate as ${\hat\sigma_{\hat\theta} = \sqrt{\frac{\sum_{j=1}^M (\hat\theta_j-\hat\theta^*)^2}{M-1}}}$ where $\hat\theta_j$ corresponds to the estimate from the $j^{{th}}$ bootstrap sample and $\hat\theta^* = \frac{\sum_{j=1}^M\hat\theta_j}{M}$ is the averaged parameter estimate. Further, the non-parametric bootstrap sampling technique allows for built-in cross validation, as the non-sampled data from the $j^{{th}}$ bootstrap sample provides us with out-of-bag (OOB) estimates for the $j^{{th}}$ model.

\cite{andrews2018addressing} notes that the conventional EM algorithm falls victim to a number of issues. The first of which is that it has the ability to converge to a local maximum, as opposed to the global maximum, in the log-likelihood resulting in an underfit. The EM may also overfit to the data, as was demonstrated in Section~\ref{sec:motex}. 

These issues can be addressed by evaluating EM algorithms over repeated bootstrap samples of the original data matrix and estimating the OOB group memberships for observations left out of the sample. Additionally, the parameter estimates are averaged over the multiple bootstrap iterations, preventing the convergence to a local maximum. It should be noted that these averaged parameter estimates are not maximum likelihood estimates; although the final likelihoods are very similar. As a result, conventional lack of progress criterion cannot be used to determine the stopping point in the algorithm. Instead, it is assumed that the algorithm has converged once the last $M$ log-likelihood values return a null from result from the Durbin-Watson test; a test for auto-correlation over a set of indices. When this is achieved, it is assumed that the algorithm has converged and deviations in the likelihood are due to the randomness from the bootstrap sampling. We denote this algorithm as BootEM, and it is presented below in Algorithm~\ref{algo:BootEM}.

\begin{algorithm}[htbp]
	\caption{Bootstrapped EM algorithm for GMM estimation}\label{algo:BootEM}
    \footnotesize
	\KwIn{$\X$}
   Initialize $\hat{\z}^{(0)}$ using either k-means or randomization\;
   Set $k=1$\;
   \While{not converged}{
        Sample with replacement from $\X$, denoting the sample as $\X^{(k)}$ and take the corresponding cluster memberships from $\z^{(k-1)}$\;
        Perform the EM Algorithm~(see Algorithm~\ref{algo:EM}) on $\X^{(k)}$ using $\z^{(k-1)}$ as cluster initializations to compute parameter estimates\;
        Compute group memberships for $\X$ using parameter estimates\;
        Check for convergence\;
    }
    \KwOut{$\hat {\theta}^*$}
\end{algorithm}

\cite{andrews2018addressing} demonstrates that this methodology addresses the issue of converging to sub-optimal solutions corresponding. However due to the bootstrapping procedure and the use of the Durbin-Watson test as a convergence criterion, it is also possible that this algorithm may take an unreasonably large amount of iterations to converge, or not converge at all~\citep{andrews2018addressing}.

\subsection{BootAECM Algorithm} \label{BootAECM}

\cite{shreeves2019bootstrap} demonstrated that the bootstrapped EM algorithm can be improved by employing factor analysis. The factor analysis model assumes that the $p$-dimensional observation $\mathbf{x}_i$ arises from
\begin{equation}
    \mathbf{x}_i = \mu + \Lambda \mathbf{u}_i + \epsilon_i,
\end{equation}
where $\mu$ is a $p$-dimensional mean vector, $\Lambda$ is a $p \times q$ matrix of factor loadings, $\mathbf{u}_i\sim N(0,\mathbf{I}_q)$ is a $q$-dimensional vector of latent variables, and $\epsilon_i \sim N(0,\Psi)$ where $\Psi$ is a $p \times p$ diagonal matrix. As a result, it can be shown that the marginal distribution for $\mathbf{x}_i$ is $\phi(\mathbf{x}_i \mid \mu,\Lambda\Lambda^\prime + \Psi)$. This methodology was applied to the field of mixture models~\citep{ghahramani1996algorithm,mclachlan2000factor} and named ``mixtures of factor analyzers.'' The reason for using this method over the conventional finite mixture model is two-fold; (1) it allows one to find underlying factors of groups and (2) the computational benefits in terms of both storage and run time are needed for moderate-to-high dimensional data sets. This is due to the fact that the number of free parameters to be estimated is reduced and the evaluation of $\hat{\Sigma}^{-1}$ and $|\hat{\Sigma}|$ are calculated in a computationally advantageous manner. Under a fully unconstrained mixture of factor analyzers model, the number of free parameters is reduced to
\begin{equation}\label{eq:freeparfac}
    G\left(p\,q-\frac{q\,(q-1)}{2}\right)+G\,p.
\end{equation}
The key difference between~\eqref{eq:freepar} and~\eqref{eq:freeparfac} is that the $p^2$ term, the main source of free parameters, no longer exists in~\eqref{eq:freeparfac}. Additionally, this implementation does not require multiple inversions of the $p \times p$ covariance matrix for the calculation of the likelihood. These calculations are instead approximated using the latent subspace, reducing computation time~\citep{mcnicholas2016mixture}. This was implemented in a bootstrap framework by \cite{shreeves2019bootstrap} using the techniques from~\cite{andrews2018addressing}, allowing for the use of the algorithm in moderate-to-high dimensional settings, which was not initially feasible with BootEM. 

\subsection{Spectral Clustering} \label{Spec}

Spectral clustering~\citep{ng2001spectral} is an extremely popular spectral methodology used in cluster analysis. Spectral clustering utilizes a singular value decomposition (SVD) of the data matrix to transform the data into a new space. Statistical analysis is then performed on the transformed data.~\cite{vempala2004spectral} show that simple spectral clustering algorithms are able to effectively learn Gaussian mixture models under minor assumptions. In our work, we perform a spectral transformation on the data matrix prior to mixture model estimation in two cases: before and after taking a bootstrap sample. We utilize the data's similarity matrix in our clustering technique, and assume that singular values along the similarity matrix's diagonal are in decreasing order. This type of spectral clustering, where one makes use of the spectrum of the data's similarity matrix to reduce its dimension, is well studied~\citep{kannan2009spectral, kumar2010Spectralclust}.

We further borrow techniques from~\cite{loffler2021optimality}. There, the authors' perform SVD on the data matrix $\X$ to perform $G$-means clustering on the matrix product $\hat \Y = \X\,V^{'}_G$, where $V^{'}_G$ is composed of the first $G$ columns of the unitary matrix $V^{'}$ from the SVD of $\X$ such that $\X = U\,\Sigma\,V^{'}$. This technique reduces the original $n\times p$ data matrix to one that is $n\times G$, significantly reducing the dimensionality of the data. By clustering on $\hat \Y$, the spectral gap condition~(\cite{von2007tutorial},~\cite{loffler2021optimality}) can be removed. Using this spectral transformation from ~\cite{loffler2021optimality}, we create an EM algorithm that fits Gaussian mixture models. We call this the Spectral-EM algorithm. While this methodology is dynamic in the sense the number of singular values is determined based on the number of expected groups present ($G$), it assumes that this information is known.

\section{Methodology} \label{method}

\subsection{Bootstrapped Clustering with Spectral Decomposition} \label{BootSpec}

We present two algorithms to address convergence to local maxima and high dimensionality using a bootstrapped EM with SVD. This is accomplished by utilizing the spectral transformation from~\cite{loffler2021optimality} and the bootstrapping procedure employed by~\cite{andrews2018addressing}, allowing for increased computational speed and reduction in overfitting to the data. These algorithms are presented in Sections~\ref{sec:alg1} and~\ref{sec:alg2}. 

\subsubsection{Spectral-BootEM Algorithm}\label{sec:alg1}

The first algorithm is an adjustment of the algorithm original developed by~\cite{andrews2018addressing}. In our adjustment, we perform an SVD on the transpose of the $n \times p$ data matrix $X$ and then perform the non-parametric bootstrap on $\hat \Y = \X\,V^{'}_G$ in addition to altering the convergence criteria. We refer to this algorithm as Spectral-BootEM, and it is given in~\ref{algo:spectBootEM}.

\begin{algorithm}[htbp]
	\caption{Spectral bootstrapped EM algorithm for GMM estimation (Spectral-BootEM)}\label{algo:spectBootEM}
    \footnotesize
	\KwIn{$\X$}
   Perform SVD on $\X$ and compute $\hat \Y = \X\,V^{'}_G$\;
   Create initial cluster memberships $\z^{(0)}$ using $\hat\Y$\;
   Set the bootstrap index $k=1$\;
   \While{not converged}{
        Sample with replacement from $\hat\Y$, denoting the sample as $\Y^{(k)}$ and take the corresponding cluster memberships from $\z^{(k-1)}$\;
        Perform the EM~(Algorithm~\ref{algo:EM}) on $\Y^{(k)}$ using $\z^{(k-1)}$ as cluster initializations to compute parameter estimates\;
        Compute group memberships for $\hat\Y$ using parameter estimates\;
        Check for convergence\;
    }
    \KwOut{$\hat {\theta}^*$}
\end{algorithm}
This algorithm will be referred to as the Spectral Bootstrapped EM, or Spectral-BootEM for short, for the remainder of this paper. 

We modify the convergence criteria from the method proposed by~\cite{andrews2018addressing}, where the Durbin-Watson test is used to check for auto-correlation. In the modification, convergence is determined by computing the component-wise relative difference in the averaged parameter between iterations space divided by the number of free parameters. Specifically, if we denote the $k^{th}$ relative difference in averaged parameter space by $R_\theta^{(k)}$, convergence is achieved when 
\begin{equation}\label{eq:conv_ineq}
    \frac{R_\theta^{(k)}}{G-1 +G\,p +G\,p\,(p+1)/2} < \epsilon_B,
\end{equation}
where $\epsilon_B$ is a small, user-defined value and not necessarily equal to the $\epsilon$ used for convergence discussed in Section~\ref{EMAlg}. The reason for doing so is simple. While the Durbin-Watson criteria has the potential of never converging to a solution (see~\cite{andrews2018addressing}), the relative difference in the averaged parameter space will reach convergence as the number of iterations increases.
Written algebraically, this takes the form 
\begin{equation} \label{eq:rel_diff}
    R_\theta^{(k)} = \sum_{g=1}^G \left\lvert \frac{\hat\pi^{*,(k)}_g - \hat\pi^{*,(k-1)}_g}{\hat\pi^{*,(k-1)}_g}\right\rvert + \sum_{g=1}^G\sum_{j=1}^p \left\lvert \frac{\hat\mu^{*,(k)}_{j,g} - \hat\mu^{*,(k-1)}_{j,g}}{\hat\mu^{*,(k-1)}_{j,g}}\right\rvert +\sum_{g=1}^G\sum_{j=1}^p\sum_{k=j}^p \left\lvert \frac{\hat\sigma^{*,(k)}_{j,k,g} - \hat\sigma^{*,(k-1)}_{j,k,g}}{\hat\sigma^{*,(k-1)}_{j,k,g}}\right\rvert .
\end{equation}
Here, $\hat\pi^{*,(k)}$, $\hat\mu^{*,(k)}$ and $\hat\sigma^{*,(k-1)}$ represent the averaged mixing proportion, mean vector, and covariance matrix, respectively, for the $k^{\text{th}}$ bootstrap sample. When computing the component wise relative differences in the covariance matrices, we do not include the lower triangles of the group matrices, as these are symmetric and hence the off-diagonal components would be contributing twice to the relative difference. In both~(\ref{eq:conv_ineq}) and~(\ref{eq:rel_diff}), we have written $p$ to represent the dimensionality of the data. However, in the case our spectral transformation $p$ is reduced down to $G$, significantly reducing the number of terms required to be calculated. We include a minimum number of bootstrap samples to be performed before we start to check for convergence.

\subsubsection{BootSpectral Algorithm}\label{sec:alg2}

The second algorithm we propose is more computationally expensive than Algorithm~\ref{algo:spectBootEM}. After creating the bootstrap sample, $ X^{(k)}$, SVD is then performed on the bootstrap sample and $\hat \Y^{(k)} = X^{(k)} V^{'}_G$ is computed. The rest of the bootstrap EM is then continued algorithm as in~\cite{andrews2018addressing}. With this method, a new SVD is computed for each sample. We note that SVD is not a unique operation. Despite this, we assume that SVD across samples is similar enough to allow for a reputable model, while also creating enough diversity during model fitting to address the issue of overfitting. The algorithm is as follows:

\begin{algorithm}[htbp]
	\caption{Bootstrapped spectral EM algorithm for GMM estimation (BootSpectral)}\label{algo:BootSpectral}
    \footnotesize
	\KwIn{$\X$}
   Perform SVD on $\X$ and compute $\hat \Y = \X\,V^{'}_G$\;
   Create initial cluster memberships $\z^{(0)}$ using $\hat\Y$\;
   Set the bootstrap index $k=1$\;
   \While{not converged}{
        Sample with replacement from $\X$, denoting the sample as $\X^{(k)}$\;
        Perform SVD on $\X$ and compute $\hat\Y^{(k)} = \X^{(k)}\,V^{'}_G$ and take the corresponding cluster memberships from $\z^{(k-1)}$\;
        Perform the EM algorithm on $\Y^{(k)}$ using $\z^{(k-1)}$ as cluster initializations to compute parameter estimates\;
        Compute group memberships for $\hat\Y$ using parameter estimates\;
        Check for convergence\;
    }
    \KwOut{$\hat {\theta}^*$}
\end{algorithm}
As in the Spectral-BootEM algorithm, lack of progress in the parameter space is again used as the convergence criteria. We call this algorithm the Bootstrapped Spectral Clustering algorithm, or BootSpectral for short. We compare this algorithm to Spectral BootEM, as well as the standard spectral clustering, BootEM, EM, and BootAECM algorithms in Section~\ref{method}.

Both the BootSpectral and Spectral-BootEM algorithms allow for the computation out-of-bag group membership estimates, which provide a more statistically favourable interpretation of the results. It assumes that an observation omitted from the bootstrap sample has not been observed, resulting in a built-in cross validation technique. 

\subsection{Potential Issues}

A bootstrap sample typically omits one third of the original data. While this allows for a natural cross-validation of the fitted model, it also poses the issue in the event that there exists a small sized group, comparative to $n$. As such, when taking a bootstrap sample there is a risk that not enough observations from potential groups are sampled, resulting either in empty groups or unrealistic parameter estimates. These are well known problems in the clustering community which many algorithms are prone to, including the ones we present in Sections~\ref{sec:alg1} and~\ref{sec:alg2}. Further, with any data transformation technique, there is a risk in losing a non-trivial amount of the variance within the original data. In data with geometrically close groups, a loss of variance due to a transformation may be unable to fit one or more groups. However, spectral transformations perform well on noisy data, so this is not a significant concern and only mentioned for completeness.

\section{Results} \label{sec:results}

\subsection{Simulations} \label{sec:sims}

\subsubsection{Mirror Data} \label{sec:mirrorSim}

We begin by returning to our mirror data set, first presented in Section~\ref{sec:motex}. While it is difficult to compare log-likelihood values between our proposed algorithms and the non-spectral algorithms due to the spectral transformation impacting model complexity, the key metrics of interest are run times, number of bootstrap samples required, and the estimated probability vectors for the centre point. The two novel algorithms are applied and compared to the EM, SpectralEM, BootEM, and BootAECM in Table~\ref{tab:mirrorRes}. The probability vector for group membership of the centre point is denoted as $\textbf{z}_c$, and the OOB group membership for the centre point is provided in the column \textit{OOB $\textbf{z}_c$}. A seed is set for reproducibility. Additional settings include  $\epsilon = 0.1$ for all EM convergence parameters, $\epsilon_B = 0.001$ for the Spectral-BootEM and BootSpectral convergence parameters, and the Durbin-Watson p-value is equal to 0.05 for the BootEM algorithm.


\begin{table}[htb]
\centering
\begin{tabular}{ lccccc }
\toprule\toprule
  & $\textbf{z}_{c}$ & OOB $\textbf{z}_{c}$  & Log-Likelihood  & Bootstraps & Elapsed Time (s) \\ 
 \midrule
 EM & $[0,1]$ & - & $-202011.6$ &  - & 0.301 \\ 
 
 AECM & $[0,1]$ & - & $-214085.7$ & - & 1.81 \\
 
 SpectralEM & $[0,1]$ & - & $-4657.0$  & - & 0.074 \\ 
 
 BootEM & $[0.490,0.510]$ & $[0.506,0.494]$ & $-202441.8$ &  4202 & 3769.097\\ 
 
 BootAECM & $[0.422,0.578]$ & $[0.475,0.525]$ & $-214244.2$ &  500 & 415.56 \\ 
 
 Spectral-BootEM & $[0.498,0.502]$ & $[0.493,0.507]$ & $-5022.308$ &  1186 & 128.447\\ 
 
 BootSpectral & $[0.503,0.497]$ & $[0.496,0.504]$ &$-5021.03$ & 2659 & 450.017\\ 
 \bottomrule\bottomrule
\end{tabular}
\caption{Comparative results for high-dimensional mirror data.}
\label{tab:mirrorRes}
\end{table}



Analyzing Table~\ref{tab:mirrorRes}, we see that the EM, AECM, and SpectralEM algorithms heavily overfit the data, as they are classifying the centre point as belonging exclusively to a single group. The BootEM algorithm performs much better at classifying the centre point, however it takes the most bootstrap samples and longest time to complete, performing at a speed of (approximately) 1.1 bootstrap samples per second. BootAECM again addresses the overfitting issue and requires significantly fewer bootstrap samples to complete than the BootEM case, however the returned probability vector is the furthest from the intuitive $[0.5,0.5]$ for the bootstrapped algorithms. It performs at a speed of 1.2 bootstrap samples per second, a slight improvement over the BootEM algorithm. Analyzing our first algorithm, Spectral-BootEM, we find that we address the issue of convergence to local maxima and overfitting in the original SpectralEM, demonstrated by the Spectral-BootEM's group membership probabilities. It required 1186 bootstrap samples and completed in 128.447 seconds, attaining a speed 9.2 bootstrap samples per second. This is a significant increase in computational speed over both the BootEM and BootAECM and is due to the spectral transformation that is performed. Our second algorithm, the BootSpectral algorithm, also addresses overfitting, achieving a similar log-likelihood as the Spectral-BootEM and nearly equivalent group membership probabilities for both the actual memberships and OOB memberships of the centre point. It performed in 2659 bootstrap samples and completed approximately 6.0 bootstrap samples per second, a decreases in speed from the Spectral-BootEM. This is due to the fact that BootSpectral is computing a unique SVD for each bootstrap sample, resulting in a more computationally expensive process. In contrast, Spectral-BootEM computes an SVD only once. This motivating example demonstrates that the novel algorithms address the issues of overfitting and convergence to local maxima solutions in a high-dimensional setting.

To further investigate the convergence and speed of our algorithm, we run the BootEM, Spectral-BootEM, and BootSpectral algorithms on the mirror data 5 times with varying seeds. All other input parameters are the same as before. The speed at which these algorithms reach convergence and the required number of bootstrap samples to do this are displayed in Table~\ref{tab:mirror_speeds}. 
\begin{table}[htb]
    
    \begin{subtable}{1\textwidth}
        \centering
        \begin{tabular}{l c c c c c}
        \toprule\toprule
        Algorithm & Run 1 & Run 2 & Run 3 & Run 4 & Run 5  \\
        \midrule
        BootEM & 1575.210 & 482.769 & 448.512 & 722.639 & 2540.199 \\ 
        Spectral-BootEM & 99.306 & 127.603 & 78.259 & 60.740 & 97.106 \\ 
        BootSpectral & 401.097 & 520.417 & 181.603 & 221.718 & 318.404 \\ 
        \bottomrule\bottomrule
       \end{tabular}
       \caption{Time (s) to reach convergence.}
       \label{tab:mirror_speeds}
    \end{subtable}
    \bigskip
    
    \begin{subtable}{1\textwidth}
        \centering
        \begin{tabular}{l c c c c c}
        \toprule\toprule
        Algorithm & Run 1 & Run 2 & Run 3 & Run 4 & Run 5  \\
        \midrule
        BootEM & 1733 & 562 & 500 & 842 & 2836 \\ 
        Spectral-BootEM & 856 & 1174 & 714 & 554 & 923 \\
        BootSpectral & 2350 & 2956 & 1182 & 1428 & 1970 \\
        \bottomrule\bottomrule
       \end{tabular}
       \caption{Bootstrap samples required for convergence.}
       \label{tab:mirror_samples}
    \end{subtable}
    \caption{Run times and samples required to reach convergence for mirror data.}
       \label{tab:mirrorSpeedAndBoots}
\end{table}

It can be seen that that the BootEM consistently takes the longest time to reach convergence, and even its quickest run is longer than the slowest run of Spectral-BootEM and four of the five runs of BootSpectral. We find that Spectral-BootEM is consistently the fastest algorithm to reach convergence its defined convergence criteria, with BootSpectral being slower than Spectral-BootEM but still much quicker than the standard BootEM. Table~\ref{tab:mirror_samples} shows that the required samples for the BootEM to reach convergence is fairly sporadic in nature, as in run 3 it reaches convergence at its minimum amount of 500 samples while run 5 takes over 2800 samples. Again, Spectral-BootEM is the most consistent of the three algorithms. Interestingly, BootSpectral in run 2 requires the most samples out of all runs for each algorithm to reach convergence. This is again due to the fact that BootSpectral recomputes the SVD for each bootstrap sample generating a larger latent space on which to estimate each model, and hence certain sample estimates may act as outliers impacting the averaged parameter space more significantly than others. Despite the longer runtime, the increased exploration can allow the algorithm to better explore the parameter space. We conclude that that our novel bootstrapped spectral clustering algorithms and convergence criteria generally produce more consistent results when compared to the BootEM.

\subsubsection{Cross-Over Data}\label{sec:cross_over}

\begin{figure}
     \centering
     \begin{subfigure}[t]{0.49\textwidth}
         \centering
         \includegraphics[width=\textwidth]{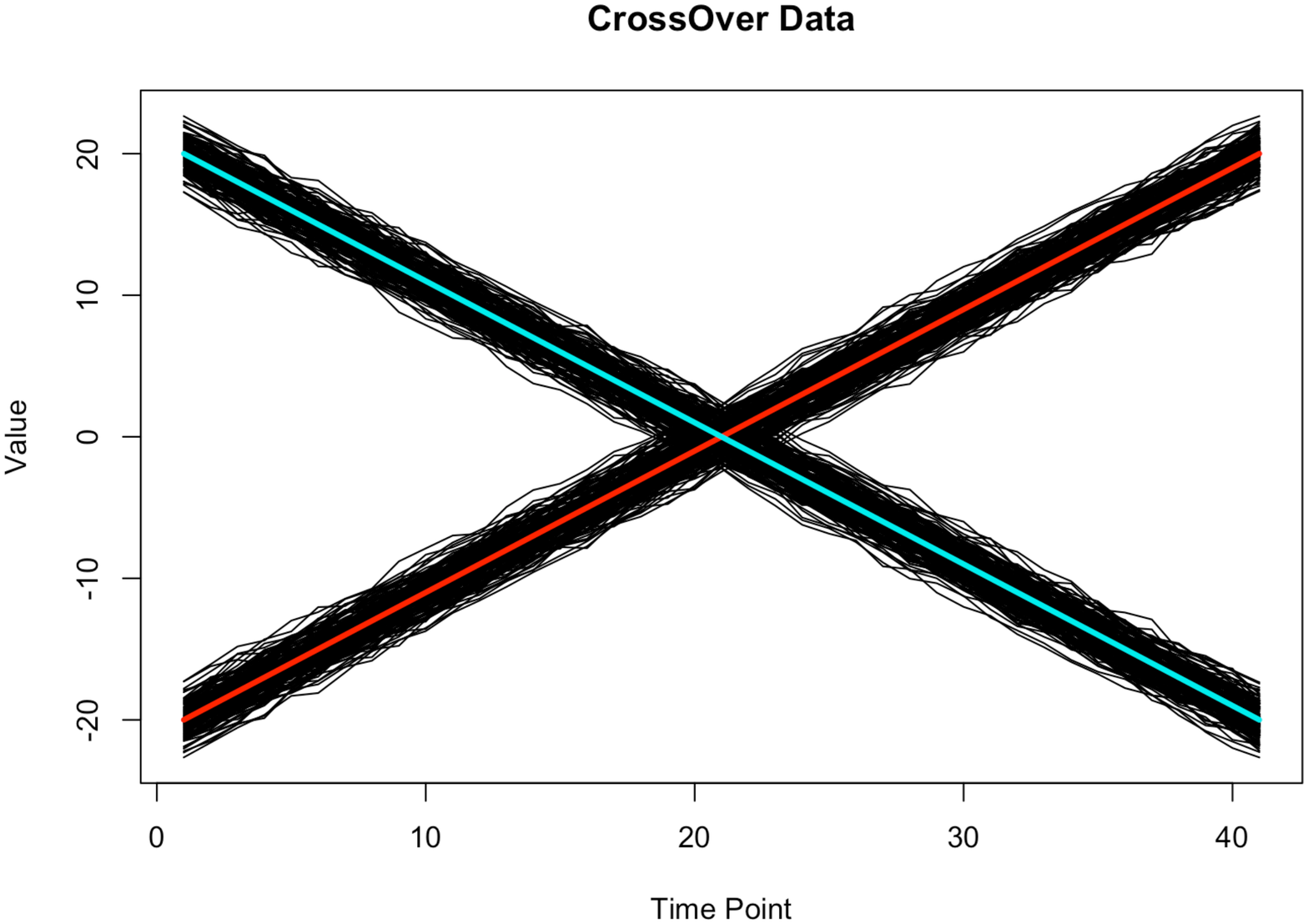}
         \caption{Cross-over data with group means in red and blue.}
         \label{fig:crossDat}
     \end{subfigure}
     \hfill
     \begin{subfigure}[t]{0.49\textwidth}
         \centering
         \includegraphics[width=\textwidth]{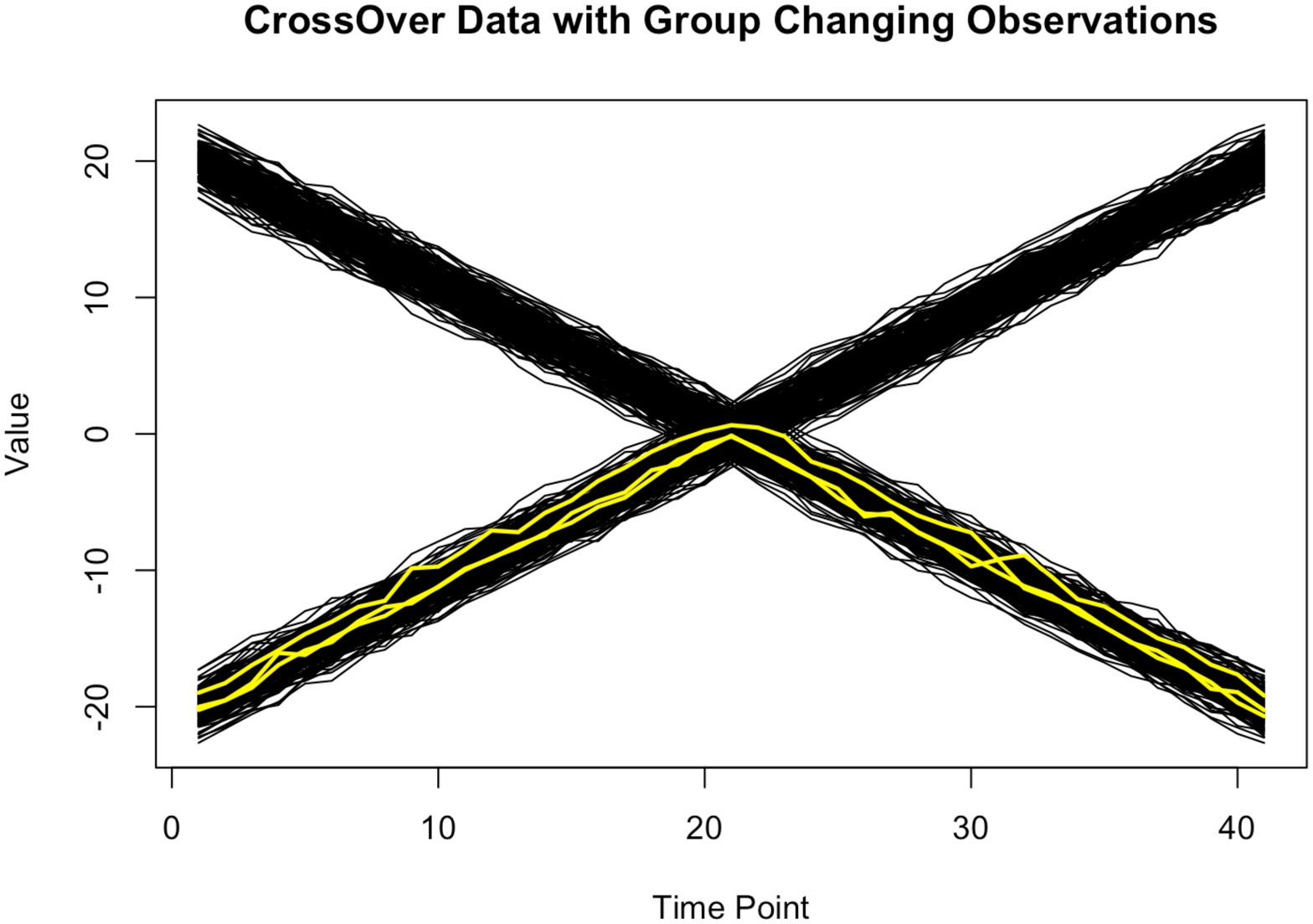}
         \caption{Cross-over data with the three group changing observations in yellow.}
         \label{fig:crossDat_groupChange}
     \end{subfigure}
        \caption{Cross-over data prior to spectral transformation.}
        \label{fig:cross_full}
\end{figure}

The next example involves a simulated longitudinal dataset. We create a two group longitudinal data set where at the halfway point, the groups cross over one another, as demonstrated in Figure~\ref{fig:crossDat}. One group trends downwards while the trends up. In this dataset, there are 150 observations in each group where at each time $t = \{1,\dots,41\}$ a value is recorded. The first group has a mean that travels from -20 to 20 by an increment of one per time step, while the second group mean travels from 20 to -20 again by an increment of one per time step. The covariance matrix for each group is constructed to be 1 on the diagonal and 0.9 on the off-diagonal. Observations are simulated using a multivariate normal distribution. Three observations are then created which belong to the bottom group for the first 20 time points, and join the other group at the cross-over event, continuing on the new path (see Figure~\ref{fig:crossDat_groupChange}). We refer to this as the cross-over data set. The EM and SpectralEM algorithm both overfit to this data, and as was seen in the mirror simulation from Section~\ref{sec:mirrorSim}, assign a group probability vector to each group changing observation of $[1,0]$. One would intuitively expect the \textit{correct} solution would be closer to a group probability vector of $[0.5, 0.5]$, as was the case with the mirror data.

\begin{figure}
     \centering
     \begin{subfigure}[t]{0.95\textwidth}
         \centering
         \includegraphics[width=\textwidth]{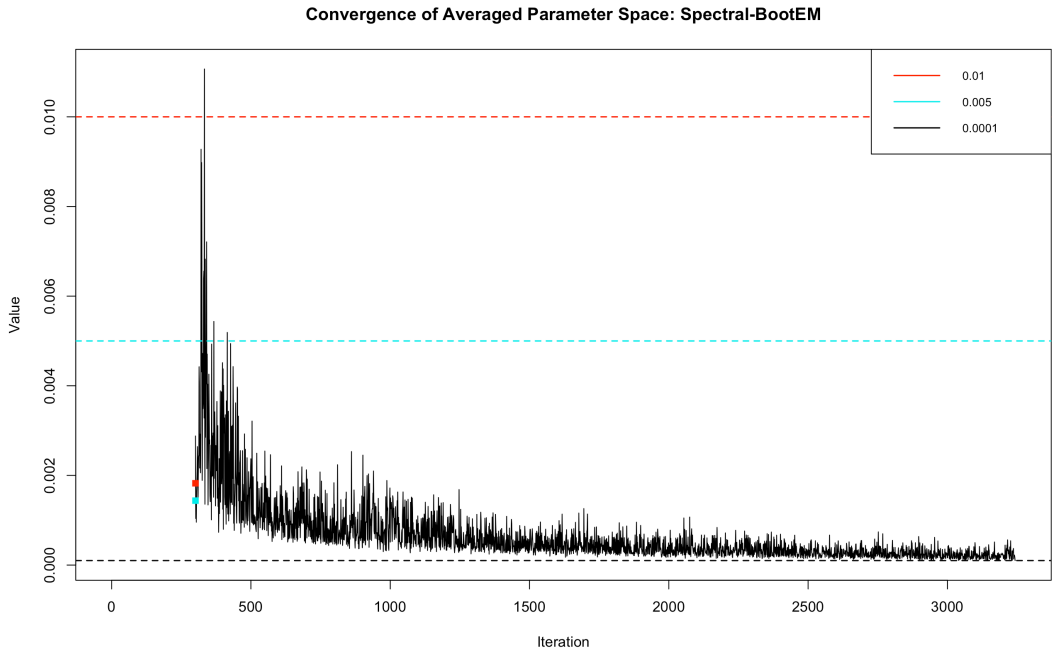}
         \caption{Impact of convergence parameter $\epsilon_B$ on convergence rate of averaged parameter space in Spectral-BootEM.}
         \label{fig:SpectBootEM_conv}
     \end{subfigure}
     \hfill
     \begin{subfigure}[t]{0.95\textwidth}
         \centering
         \includegraphics[width=\textwidth]{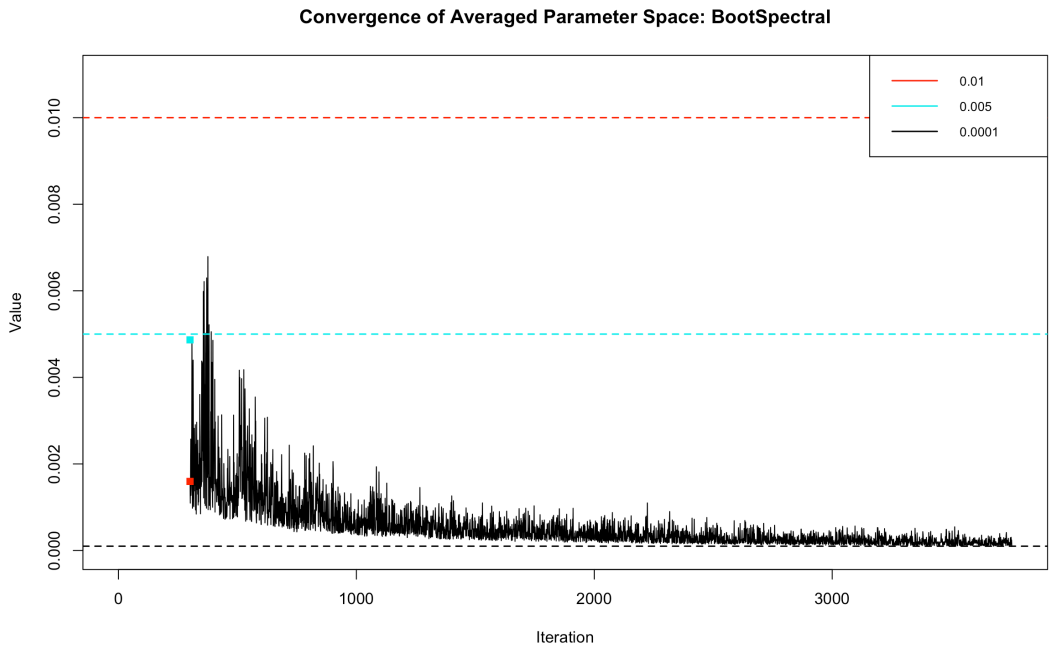}
         \caption{Impact of $\epsilon_B$ on convergence rate of averaged parameter space in BootSpectral.}
         \label{fig:BootSpect_conv}
     \end{subfigure}
        \caption{Convergence of bootstrapped spectral algorithms with varying $\epsilon_B$.}
        \label{fig:epsilon_spectral_conv}
\end{figure}

We run three experiments to demonstrate the versatility of our convergence criteria, and compare it to the results of the BootEM algorithm. For both the Spectral-BootEM and BootSpectral, the model is fitted using the cross-over data set with $\epsilon_B$ set to 0.01, 0.005, and 0.0001. The resulting group probability vectors for the group changing observations are analyzed in Tables~\ref{tab:crossOver_eps005},~\ref{tab:crossOver_eps0005}, and ~\ref{tab:crossOver_eps0001}. For all cases, we set the minimum bootstrap samples to be 300. The group changing observations are denoted as \textit{Obs 1}, \textit{Obs 2}, and \textit{Obs 3}. We further analyze the convergence of the averaged parameter space of Spectral-BootEM and BootSpectral in Figures~\ref{fig:SpectBootEM_conv} and~\ref{fig:BootSpect_conv}, respectively.

\begin{table}[htb]
    
    \begin{subtable}{1\textwidth}
        \centering
        \begin{tabular}{ lccc }
        \toprule\toprule
          & Spectral-BootEM $\textbf{z}_{i}$  & BootSpectral $\textbf{z}_{i}$  & BootEM  $\textbf{z}_{i}$\\ 
         \midrule
         Obs 1 & $[0.703,0.297]$ & $[0.620,0.380]$ & $[0.493,0.507]$ \\ 
         
         Obs 2 & $[0.710,0.290]$ & $[0.617,0.383]$ & $[0.446,0.554]$ \\ 
         
         Obs 3 & $[0.707,0.293]$ & $[0.620,0.380]$ & $[0.490,0.510]$ \\ 
         \bottomrule\bottomrule
        \end{tabular}
        \caption{$\epsilon_B = 0.01$.}
        \label{tab:crossOver_eps005}
    \end{subtable}
    \medskip
    
    \begin{subtable}{1\textwidth}
        \centering
        \begin{tabular}{ lccc }
        \toprule\toprule
          & Spectral-BootEM $\textbf{z}_{i}$  & BootSpectral $\textbf{z}_{i}$  & BootEM  $\textbf{z}_{i}$\\ 
         \midrule
         Obs 1 & $[0.490,0.510]$ & $[0.577,0.423]$ & $[0.493,0.507]$ \\ 
         
         Obs 2 & $[0.480,0.520]$ & $[0.573,0.427]$ & $[0.490,0.510]$ \\ 
         
         Obs 3 & $[0.487,0.513]$ & $[0.573,0.427]$ & $[0.490,0.510]$ \\ 
         \bottomrule\bottomrule
        \end{tabular}
        \caption{$\epsilon_B = 0.005$.}
        \label{tab:crossOver_eps0005}
    \end{subtable}
    \medskip

    \begin{subtable}{1\textwidth}
        \centering
        \begin{tabular}{ lccc }
        \toprule\toprule
          & Spectral-BootEM $\textbf{z}_{i}$  & BootSpectral $\textbf{z}_{i}$  & BootEM  $\textbf{z}_{i}$\\ 
         \midrule
         Obs 1 & $[0.487,0.513]$ & $[0.509,0.491]$ & $[0.493,0.507]$ \\ 
         
         Obs 2 & $[0.485,0.515]$ & $[0.506,0.494]$ & $[0.490,0.510]$ \\ 
         
         Obs 3 & $[0.486,0.514]$ & $[0.508,0.492]$ & $[0.490,0.510]$ \\ 
         \bottomrule\bottomrule
        \end{tabular}
        \caption{$\epsilon_B = 0.0001$.}
        \label{tab:crossOver_eps0001}
    \end{subtable}

    \caption{Averaged \textbf{z} vectors for group changing observations.}
    \label{tab:epsilon_convs}
\end{table}

Analyzing first Table~\ref{tab:crossOver_eps005}, where $\epsilon_B = 0.01,$ we see that the Spectral-BootEM still partially overfits to the data, as is shown by the poor probabilities in the group probability vectors. Meanwhile, the BootSpectral algorithm performs well, with group probability vectors close to $[0.5,0.5]$. The BootEM also performs admirably, and much better than Spectral-BootEM. However, as our convergence criteria allows for flexibility in $\epsilon_B$, we lower this to $0.005$, and display these results in table~\ref{tab:crossOver_eps005}. Immediately we see a an improvement in membership probabilities in both the Spectral-BootEM and BootSpectral algorithms. Note that the BootEM remains unchanged, as it's convergence criteria uses a Durbin-Watson test for auto-correlation, and as such does not allow for any flexibility assuming a significance level of $\alpha = 0.05$ is kept. 

Next, the convergence criteria is again reduced such that the relative difference (scaled by the number of free parameters) in the average parameter space cannot exceed 0.0001, and display the results in Table~\ref{tab:crossOver_eps0001}. Here, we find the most intuitively logical results for Spectral-BootEM and BootSpectral, as both provide group probability vectors close to $[0.5,0.5]$ for all changing observations. This demonstrates the flexibility of the convergence criteria. 

We now analyze the behaviour of convergence criteria for the cross-over data. Beginning with Spectral-BootEM in Figure~\ref{fig:SpectBootEM_conv}, we see that setting $\epsilon_B = 0.0001$ results in the most iterations, as is expected, and achieves an asymptotic behaviour demonstrating that bootstrap samples will have a smaller impact on the overall averaged parameter space; something which has been found to have a significant negative impact the convergence BootEM algorithm~\citep{andrews2018addressing}. The results are analogous for the BootSpectral, which we show in Figure~\ref{fig:BootSpect_conv}. In both figures, a large $\epsilon_B$ results in extremely fast convergence. As we do not begin checking for convergence before iteration 300, it is possible that our algorithms could converge earlier for these less strict values of $\epsilon_B$. To avoid potential overfits we maintain that a minimum of 300 bootstraps samples is required. We note that in Figure~\ref{fig:SpectBootEM_conv}, despite the convergence measure being below 0.005 for multiple iterations, extreme bootstrap samples significantly impact the averaged parameter space, and we see a large spike occurring early in the convergence measure growing to larger than 0.01. This is despite the criterion being relatively small to begin with, as our two other runs with larger $\epsilon_B$ values had already converged. As such, we find the choice of $\epsilon_B$ to be of great importance to achieve a stable averaged parameter space. These figures also show the importance of suitable minimum iteration count, in order to avoid any significant spikes in the relative difference of the averaged parameter space between iterations in the early stages of the bootstrapped model estimation.

\subsection{Real Data}\label{realDat}

\subsubsection{Raman Data}

Raman spectroscopy~\citep{mccreery2005raman} is an optical interrogation method which consists of using inelastic light scattering to identify constituent chemicals by their vibrational modes. The acquired spectra can therefore provide detailed information on a range of constituents within a single sample acquisition. Changes in the positioning of peaks, as well as their relative amplitudes, can be used to identify the change in molecular dynamics due to a particular perturbation of a system. 

\cite{matthews2015radiation} collected a series of Raman spectra that are classified into three different groups; lung (H460), breast (MCF-7), and prostate (LNCaP) tumor cells. Their research identified that using dimensionality reduction techniques, such as principal component analysis, is a potential way to identify the discrepancies between different cell types while also monitoring the underlying biochemical changes within given cell lines. Specifically, they found that one can distinguish the lung and breast tumor cells from the prostate tumor cells through the levels of glycogen found in their underlying spectra. \cite{deng2020monitor} further verified this using a variety of dimensionality reduction techniques, such as nonnegative matrix factorization and nonnegative least squares, comparing the results to those previously found by \cite{matthews2015radiation}. Additionally, they were able to identify both glycogen and lipid-like components to further discern the disparity between different cell lines. Herein, we elect to utilize our novel methodology to identify the underlying groups structure under the assumption that there is no group information available.

\begin{figure}
\centering
    \includegraphics[width=0.95\textwidth]{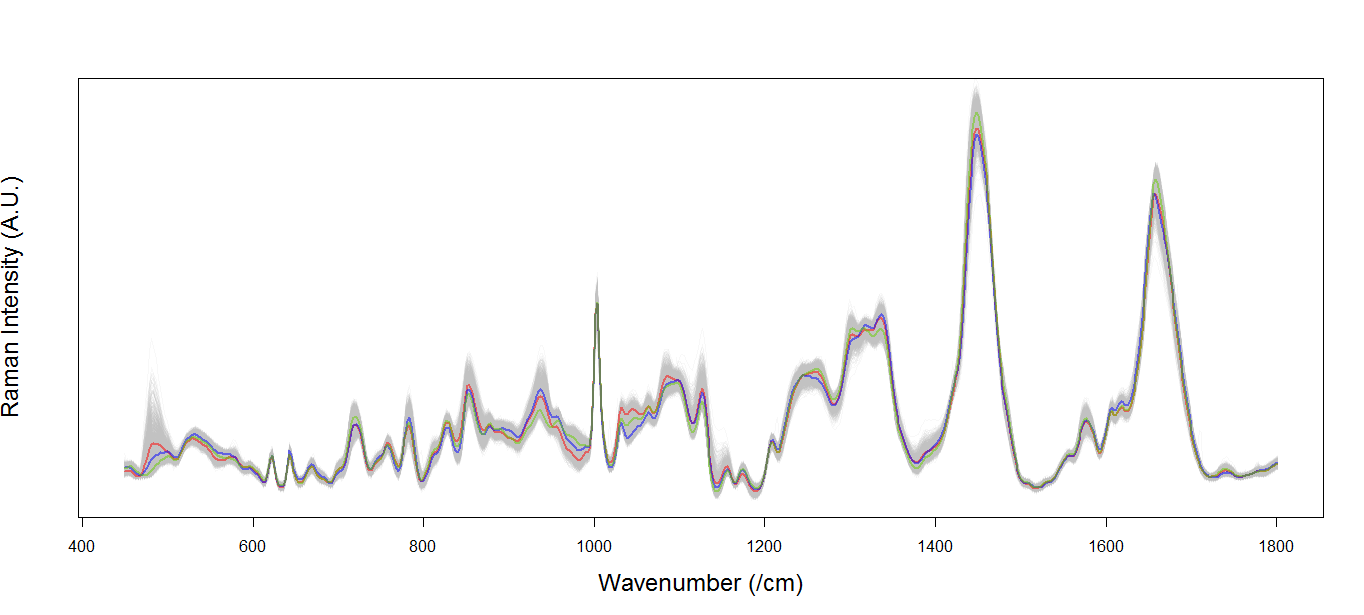}
    \caption{A demonstration of observations in the Raman spectroscopy data set. Observations are coloured in grey, whereas the group means are displayed in red, green, and blue.}
    \label{fig:Raman}
\end{figure}

The data consists of 1080 observations in the 381-dimensional space. Here each observation consists of a series of Raman intensities (arbitrary units) at specific wavenumbers ($cm^{-1}$). Figure~\ref{fig:Raman} displays the collection of observations in our data set coloured in grey. Additionally, we include the group means which identify the averaged spectra for the H460 (red), MCF-7 (blue), and LNCaP (green) cell lines. Each group has an even number (360) of observations, which are differentiated by radiation dose, number of days, and individual number; all of which are left out of the clustering model.


\begin{table}[htb]
\centering
\begin{tabular}{ lccccc }
\toprule\toprule
  Algorithm & CR & ARI & Log-Likelihood  & Bootstraps & Elapsed Time (s) \\ 
  \midrule
 AECM & 1.00 & 1.00 & 3,648,714 & - & 2000.84 \\ 
 
 SpectralEM & 0.99 & 0.98 & 21,270.54 & - & 0.83 \\ 
 
 Spectral-BootEM & 0.99 & 0.98 & 21,250.56 & 300 & 53.75 \\ 
 
 BootSpectral & 0.99 & 0.97 & 20,874.64 & 300 & 195.55 \\ 
 \bottomrule\bottomrule
\end{tabular}
\caption{Comparative results for Raman spectroscopy data. Classification rate (CR) and adjusted Rand index (ARI) are both measures of agreement between true classes and the assigned cluster memberships.}
\label{tab:ramanRes}
\end{table}

Table~\ref{tab:ramanRes} displays results from the Spectral-EM algorithm and its bootstrapped counterparts on the Raman spectroscopy data set. These results include the percentage of correctly classified observations (CR), the adjusted rand indices (ARI; \cite{hubert1985comparing}), log-likelihoods, number of bootstrap iterations, and elapsed run time of the algorithms. While the conventional Spectral-EM performs well, it should be noted that both the Spectral-BootEM and BootSpectral algorithms also achieve similar results. This is indicative of a clear solution in the data. However, it should be noted that this clustering problem is not feasible using the conventional EM algorithm, its bootstrapped counterpart, or the BootAECM due to the requirement of $p \times p$ matrix inversions; an operation that often results in computational errors. This is due to the fact that matrix singularities arise from a deficient number of observations within groups and is further compounded when taking into consideration the data left out in a bootstrapped sample. While the conventional AECM algorithm does not require a $p \times p$ matrix inversions, its bootstrapped version does since it requires an inversion of the averaged sample space covariance matrix. This cannot be avoided as an inversion using the averaged latent parameters can be perturbed due to discoveries of different orthogonal rotations in the data. The conventional AECM algorithm performs admirably, with a perfect classification rate and ARI. However, its elapsed run time is over ten times the next slowest algorithm in table~\ref{tab:ramanRes}. While it is the user's decision to decide which algorithm is more appropriate, the expense of a few misclassifications in exchange for much faster run time was deemed worthy in this case. Further, the using the bootstrapped algorithms assuage concern of overfitting to the data, something that is a more likely in AECM algorithm given its perfect classification rate.

\section{Conclusion} \label{conc}

We have demonstrated that our two novel bootstrapped spectral clustering algorithms effectively and efficiently address the previously defined issues of convergence to sub-optimal solutions. These algorithms have shown to achieve a significant boost in speed without a loss in accuracy occurring from the spectral transformation. We demonstrated the versatility and efficacy of our novel algorithms compared to two prior bootstrapped clustering algorithms on simulated and real data experiments. Our choice of convergence technique by way of the relative difference in the averaged parameter space allows for a more consistent model solution than other convergence criteria methods, such as the Durbin-Watson test.

Both algorithms have their benefits and drawbacks. The Spectral-BootEM is easier to interpret and less likely to result in substantial deviations in the averaged parameter space for each bootstrapped model, while the BootSpectral is more computationally expensive but allows for the discovery of latent spaces that may not be immediately obvious in the full data set. This results in greater variation of the fitted models of each bootstrap, causing full exploration of the parameter space. 


\section{Acknowledgements} \label{Acknowledgements}
The authors kindly thank Dr.~Jeffrey Andrews for providing infrastructure allowing for a more streamlined research process. The authors also thank Dr.~Jeffrey Andrews and Dr.~Ryan Browne for permission to use the cross-over data simulation example. Finally, we thank Dr.~Andrew Jirasek and his medical physics research group for permission to use the Raman Spectroscopy data.

\newpage
\bibliographystyle{apalike}
\bibliography{refs}

\end{document}